\documentclass[runningheads]{llncs}
\usepackage[T1]{fontenc}
\usepackage{subcaption}
\usepackage{graphicx}
\usepackage{multirow}
\usepackage{amssymb}
\usepackage{dsfont}
\usepackage{xcolor}
\usepackage{float}
\usepackage{hyperref}
\hypersetup{
    colorlinks=true,
    linkcolor=red,
    filecolor=magenta,      
    urlcolor=blue,
    pdftitle={Overleaf Example},
    pdfpagemode=FullScreen,
    }

\usepackage{color}

\begin{document}
\title{Gaining Explainability from a CNN for Stereotype Detection Based on Mice Stopping Behavior}
\titlerunning{Explainability from a CNN for Stereotype Detection}

\author{Raul Alfredo de Sousa Silva\inst{1}\and
Yasmine Belaidouni\inst{2}\and
Rabah Iguernaissi\inst{1} \and
Djamal Merad\inst{1} \and
Séverine Dubuisson\inst{1}}

\authorrunning{de Sousa Silva et al.}

\institute{Laboratoire d'Informatique et des Systèmes, CNRS, Aix-Marseille University, Marseille, France\\
\email{\{raul-alfredo.de-sousa-silva, rabah.iguernaissi, djamal.merad, severine.dubuisson\}@lis-lab.fr}\\
\url{https://im.lis-lab.fr/} \and
Aix-Marseille University, CNRS, IBDM, UMR7288, Marseille, France\\
\email{yasmine.belaidouni@univ-amu.fr}}

\maketitle              % typeset the header of the contribution
\begin{abstract}
Understanding the behavior of laboratory animals is a key to find answers about diseases and neurodevelopmental disorders that also affects humans. One behavior of interest is the stopping, as it correlates with exploration, feeding and sleeping habits of individuals. 
To improve comprehension of animal's behavior, we focus on identifying trait revealing age/sex of mice through the series of stopping spots of each individual. We track 4 mice using LiveMouseTracker (LMT) system during 3 days. Then, we build a stack of 2D histograms of the stop positions. This stack of histograms passes through a shallow CNN architecture to classify mice in terms of age and sex. 
We observe that female mice show more recognizable behavioral patterns, reaching a classification accuracy of more than 90\%, while males, which do not present as many distinguishable patterns, reach an accuracy of 62.5\%. 
To gain explainability from the model, we look at the activation function of the convolutional layers and found that some regions of the cage are preferentially explored by females. Males, especially juveniles, present behavior patterns that oscillate between juvenile female and adult male.


\end{abstract}

\section{Introduction}
\label{sec:intro}

Automated Animal Behavior Analysis has become an important active research field during the last decade. It has been furnished with increasingly powerful algorithms by computer vision and machine learning. With the possibility of recording experiments with laboratory animals, the manual analysis became obsolete and unfeasible for a human, once the amount of data is beyond the capacity of processing for an experimenter. In special, manual annotations are prone to experimenter bias and drift throughout the annotation process. Thus, researchers have proposed algorithmic solutions to tackle a number of tasks in the field. The animals studied varied from ants \cite{gal_antrax_2020} to fruit flies \cite{karashchuk_anipose_2021}, mice  \cite{le_markerless_2024,mathis_deeplabcut_2018}, zebrafish \cite{xu2017lie}, and many others.

Understanding the behavior of laboratory animals is crucial to the discovery of disease mechanisms, neural correlations to behaviors, and efficient treatments, which can be later tested on humans. That is why it is important to have consistent analysis protocols and metrics.

In general, experiments are designed to have a minimal population of animals that ensures statistical significance for tests, but this population is most often not enough to train a deep learning model for detecting differences between individuals. Thus, automatic methods capable of reliably highlighting behavioral differences would represent a new step towards understanding stereotypes in neuroscience.
Here, we borrow the definition of stereotype from Battle, which says ``a stereotype is a held popular belief about specific social groups or types of individuals'' \cite{battle_chapter_2012}. In this case, sex and age are characteristics separating those ``types of individuals''. 

Although many proposed algorithmic solutions tackle different problems in fields such as tracking, identity recognition and behavior detection, the recognition of impaired and stereotyped behaviors is still not fully explored. 
In this work, we propose a pipeline to recognize mice age and sex, using their series of stopping points. 

The stopping spots of the mice are related to exploration, food and water consumption and sleeping patterns. Our hypothesis is that distinct stereotypes can be classified based on the temporal sequence of stops, because of the different dominance strategies employed by individuals in each gender and age, which affects exploration, feeding, and sleeping behaviors of individuals. 

Using the Live Mouse Tracker (LMT) tracking system \cite{de2019real}, we recorded for multiple cages the position and identity, as well as a set of behaviors of four house-grouped mice during 3 days. Then, with a concatenation of single layers Convolutional Networks, we can show that stopping spot sequences actually store relevant information about animal characteristics and can be sufficient to predict the age and sex of individuals.

The following sections are structured as follows: examination of the related work (Section \ref{sec:sota}), description of the proposed model (Section \ref{sec:methods}), presentation of the experiments and results (Section \ref{sec:experiments}) and a conclusion (Section \ref{sec:conclusion}).

\section{Related Work}
\label{sec:sota}

A variety of approaches  were already proposed to distinguish animals of different sexes, ages or disease conditions, mostly based on social interactions. Mice are particularly studied due to the extensive genetic knowledge already discovered by researchers and their similarity with human \cite{perlman2016mouse}. Some works were presented with the goal of detecting behaviors in mice, without considering behaviors that could only appear in one stereotype or the change in the way of expressing those behaviors due to stereotypical variations. Burgos-Artizzu \textit{et al.} \cite{burgos2012social} proposed a classifier to categorize behaviors on a dataset of mice in resident-intruder interactions. The approach is based on spatio-temporal energy and agent trajectories that are used to classify behavior into 13 classes using AdaBoost classifiers. York \textit{et al.} \cite{york_flexible_2020} on their turn, presented a pipeline based on postural and displacement features to predict a sequence of behaviors and find neural correlations with movement patterns. 

Other works relied on behaviors that could change between stereotypes, in particular sex, but only measured statistical differences at the population level without deepening the discussion about how to identify, at the individual level, the stereotype. 
Netser \textit{et al.} \cite{netser_novel_2017}, for example, presented a new tracking system for analysis of social preference dynamics, and highlight that a loss of social preference is observable in females. The time passed investigating the novel individual reduces quickly over time for females, while males keep investigating the new mouse up to 5 minutes after introduction.
Zilkha \textit{et al.} \cite{zilkha_sex-dependent_2023} studied the behavioral personalities in male and female mice cages to conclude that, although males and females both establish dominance, each sex has its own strategies. Clein \textit{et al.} \cite{clein_automated_2024} investigated dominance hierarchies in mixed-sex groups of mice. Through the detection of sequences of behaviors, the authors showed that the submissive mouse has a mechanism to de-escalate social conflicts. We can also find in the literature multiple works where significant statistical difference was found between groups of tested animals, either due to  specific gene variation \cite{ferhat_excessive_2023,huzard_impact_2022}, or exposition to drugs \cite{maisterrena_female_2024}.

Despite these findings, none of these works tried to infer individuals' sex, strain or hierarchy, based on the extracted features. Neither of them intended to unveil distinctive features at the individual level that could identify patterns in the animals related to any of the stereotypes studied. In this work, we propose an explainable neural network model to identify stereotypes in individuals through the series of stopping spots of each animal.

\section{Stack of Histograms Feature Space}
\label{sec:methods}

We recorded 24 male and 24 female mice for 3 days. Each mouse was recorded at two different ages. The first recording happened around 40 days of life, and the second recording happened when mice were about 6 months. We have, then, 96 recordings in total. Animals were housed in groups of four. The recordings were done into a 50x50 cm cage with food and water \textit{ad libitum}. The arrangement of objects in the cage was the same for all experiments. A system called Live Mouse Tracker (LMT) \cite{de2019real} was used to manage recordings and to retrieve positions (center of mass) of mice into the cage at 30 fps. Based on these stopping coordinates and on the definition of stop as the mice velocity being $\leq$ 5 cm/s, we retrieve all stopping spots in which the mouse remained for at least 1 s. This allowed us to have stop detections consistent with what is seen in video recordings.

\begin{figure}[H]
    \centering
    \begin{subfigure}{\textwidth}
        \includegraphics[width=\textwidth]{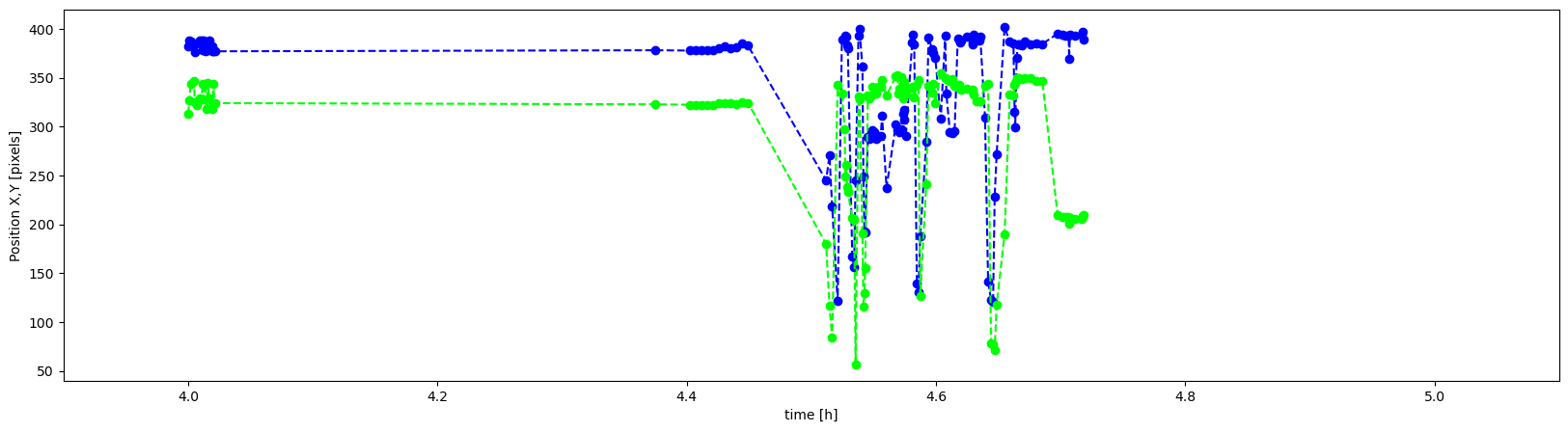}
        \caption{Coordinates through time. Blue are x coordinates, and green the y coordinates. Markers indicate the initial time of the stopping period.}
        \label{fig:irregular_ts}
    \end{subfigure}
    
    \begin{subfigure}[t]{0.49\textwidth}
        \includegraphics[width=\textwidth]{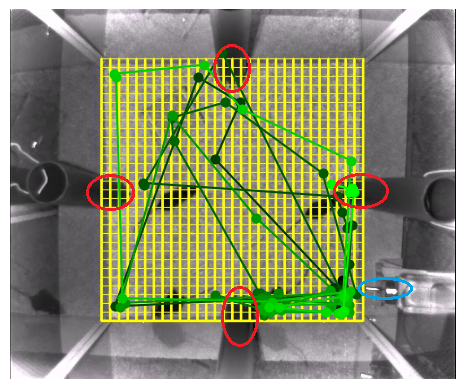}
        \caption{Stop spots over time. The clearer the marker, the further in time. The yellow grid shown is the grid created over the cage coordinates. Feeders are in red, water spot in blue.}
        \label{fig:cage}
    \end{subfigure}
    \hfill
    \begin{subfigure}[t]{0.49\textwidth}
        \includegraphics[scale=0.5]{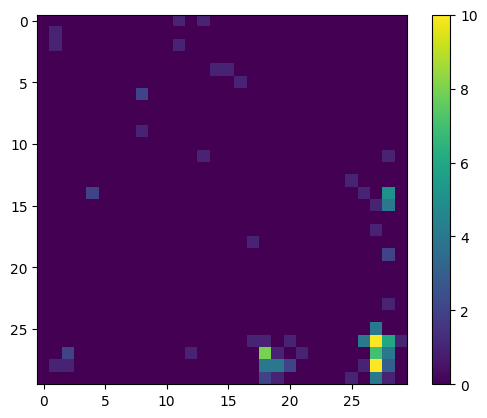}
        \caption{Histogram map of the stop spots. The lighter, the higher is the number of stops in this spot. Here, the mouse spent most of its time next to the water spot.}
        \label{fig:histogram}
    \end{subfigure}
    \caption{An example of a 1-hour time series of stops of one mouse. The most often visited spots are the nest and the feeders.}
    \label{fig:data}
\end{figure}

The set of stops can be modeled as an irregular time series, because the event sequences have different lengths among mice, and they happen at irregular intervals. This is illustrated by Figure \ref{fig:irregular_ts}. In order to have regularly sampled data and the same dimension for every recording, we compute a stack of 2D histograms of the stopping spots. Firstly, the cage surface is split into a regular grid of size $(N\times N)$, as shown in figure \ref{fig:cage}. Also, the stopping coordinates sequences are split in $T$ regular intervals. Then, we count the number of stop events $\mathbb{S}^{t} \in \mathds{R}^{N\times N}$ happening in a tile $\{b_x, b_y\}=\{\left[1\ ...\ N\right], \left[1\ ...\ N\right]\}$ of the grid for each interval $t = \left[ 1\ ...\ T\right]$. Thus, we have a stack of histograms $\mathbf{S} = [\mathbb{S}^{1}; ...;  \mathbb{S}^{T}] \in \mathds{R}^{T \times N \times N}$ for each recording. These histograms are the features used to distinguish animals and to detect stereotypes. They represent the evolution in time of the spots' preferences of each individual throughout the experiment. We process these sets of histograms as stacked 2D maps, characterizing the behavior of the animal. Each map in the time dimension is treated as a channel of an image. An example of such maps is shown in Figure \ref{fig:histogram}.

\begin{figure}
    \centering
    \includegraphics[scale=0.4]{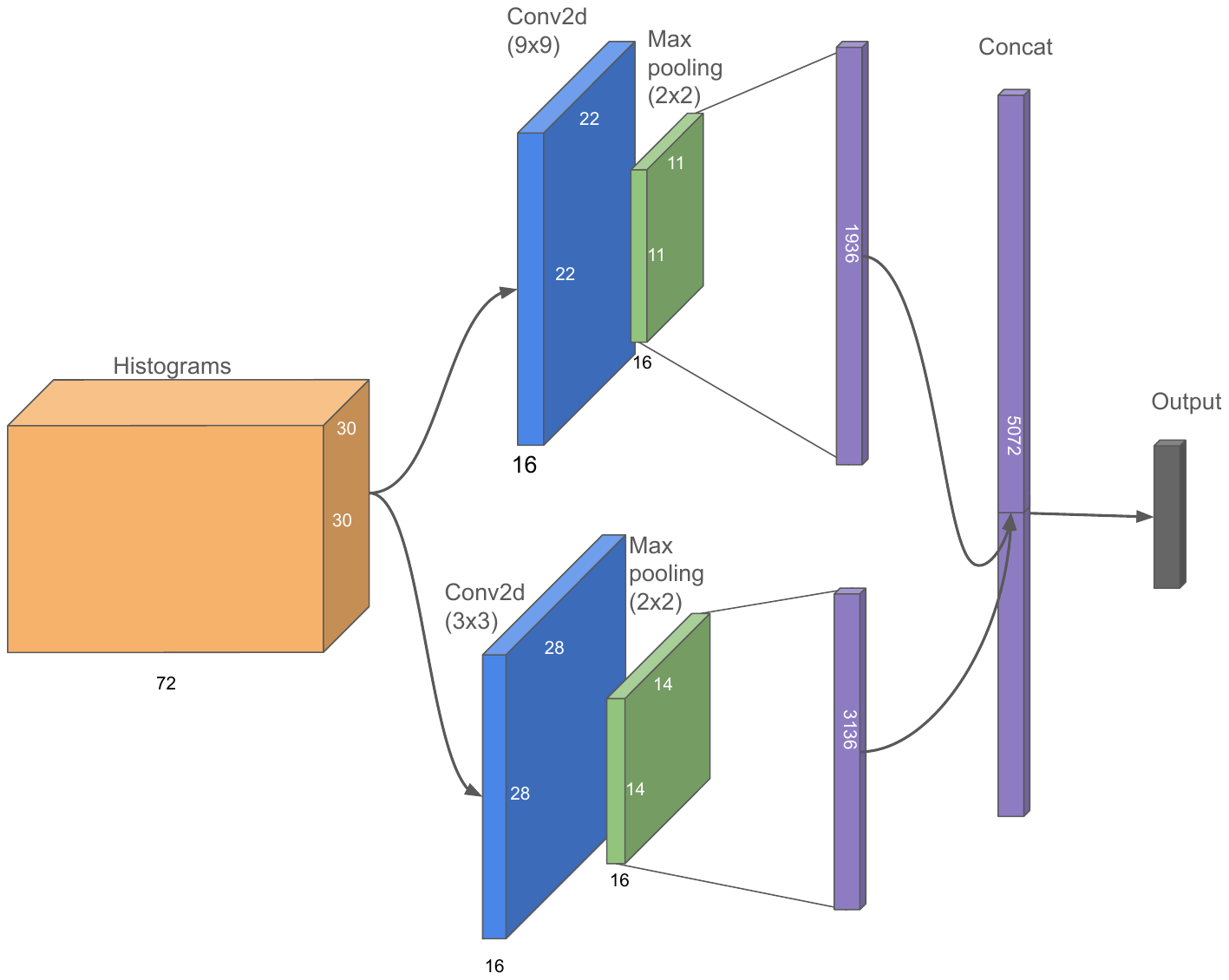}
    \caption{Proposed CNN architecture to predict sex and age from histogram maps of mice.}
    \label{fig:cnn}
\end{figure}

We build a simple CNN architecture. It is composed of 2 convolutional layers receiving as input the histogram maps. The channels of both layers are flattened and concatenated to pass through a single linear layer. Figure \ref{fig:cnn} illustrates the chosen architecture. The choice of such a network has two goals: avoiding overfitting due to the reduced size of the dataset, and extracting the explainability of the model choices through the convolution kernel outputs assessment. We train the model to classify data into 4 classes: \texttt{adult-males}(AM), \texttt{juvenile-males}(JM), \texttt{adult-females}(AF) and \texttt{juvenile-females}(JF).

\vspace{-20pt}

\section{Experiments}
\label{sec:experiments}

For the construction of the histogram maps, stopping spots series were split into $T=72$ 1-hour intervals. The number of subdivisions in each dimension of the cage was $N=30$. Our architecture was composed of two convolutional layers with ReLU as activation function. Each layer has different kernel sizes of ($3\times 3$) and ($9\times 9$), both with 16 channels. Output of both layers are subsampled with a Max-pooling layer and then, flattened and concatenated to be sent to a single linear layer for classification. Learning was done with a Cross Entropy Loss, using an SGD optimizer, and learning rate was set to $1e^{-3}$. 

As we observed that maps from mice of the same cage had strong common traits, we adopted a more challenging learning strategy: dataset is separated per cages and trained using a leave-one(cage)-out cross validation strategy. This avoids the model to rely on common cage features to recognize gender or age, forcing it to learn from generalizing the dynamic features of other cages.

We see the classification results in Table \ref{tab:4class}. Looking at the table, we observe that \texttt{juvenile-males} are misclassified as \texttt{juvenile-females}, meaning that they share some patterns.
We also realize that the model struggles to separate \texttt{adult-males} and \texttt{juvenile-males} mice. This seems an indicative that male mice take less time to transit from juvenile age to adulthood, which justifies the same displacement patterns at both ages. On the other hand, females had a distinguishable shift in behavior between the two ages. 

\begin{table}
    \centering
    \begin{tabular}{ccr|r|r|r}
    \multicolumn{2}{c}{}        & \multicolumn{4}{c}{Prediction} \\ 
                           &    & AM & JM & AF & JF \\ \cline{3-6}
    \multirow{5}{*}{Labels}& AM & 22 &  2 &  0 &  0 \\ \cline{3-6}
                           & JM &  9 &  8 &  1 &  6 \\ \cline{3-6}
                           & AF &  0 &  4 & 20 &  0 \\ \cline{3-6}
                           & JF &  0 &  0 &  0 & 24 \\ 
    \end{tabular}
    \caption{Confusion matrix of sex-age classification. Average accuracy is 77.1\%, but accuracy is better among females (91.7\%) than males (62.5\%).}
    \label{tab:4class}
\end{table}

\vspace{-30pt}

In order to have a baseline for comparison, we tested k-Nearest Neighbors and SVM classifiers on PCA reduced data. We applied the dimensionality reduction (PCA) over the flattened stack of histograms (D=72*30*30=64800) and kept the first 3 dimensions. The 1-Nearest Neighbor resulted in an accuracy of 0.62, while SVM resulted in an accuracy of 0.72, all of them, considerably lower than the accuracy obtained with the CNN (0.77).

\begin{figure}
    \centering
    \begin{subfigure}{0.49\textwidth}
        \includegraphics[width=\textwidth]{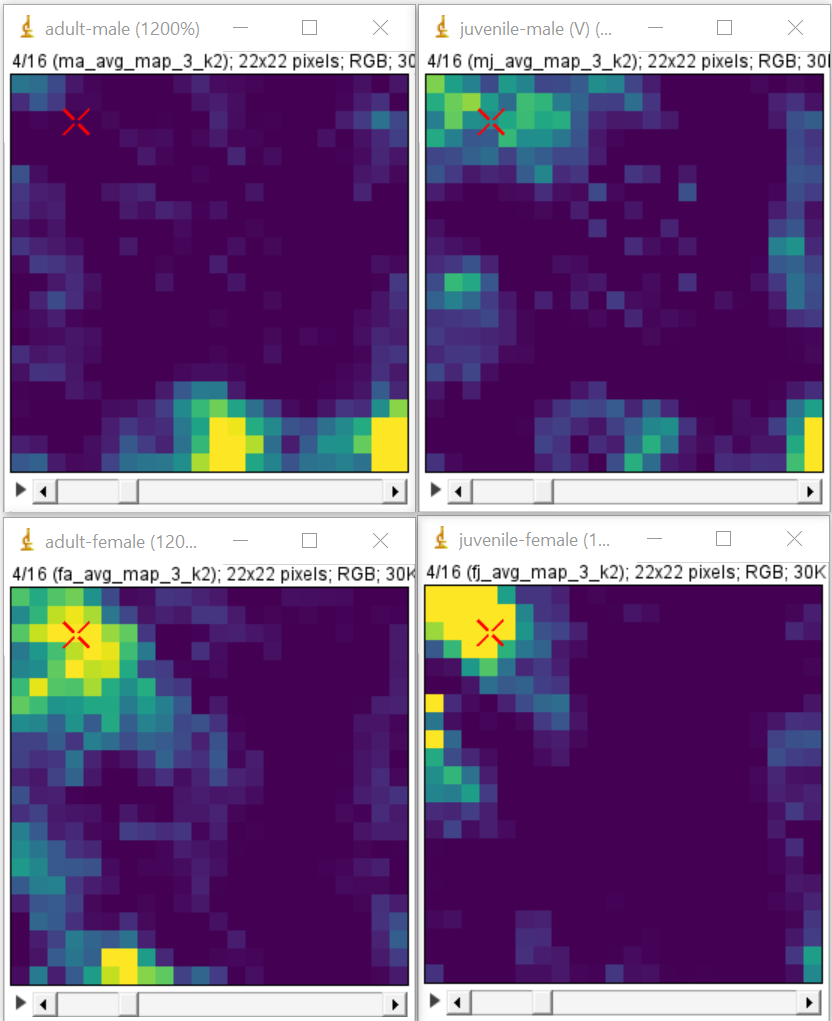}
        \caption{Output of channel 4 of second convolutional layer (9x9 kernel).}
        \label{fig:maps3}
    \end{subfigure}
    \begin{subfigure}{0.49\textwidth}
        \includegraphics[width=\textwidth]{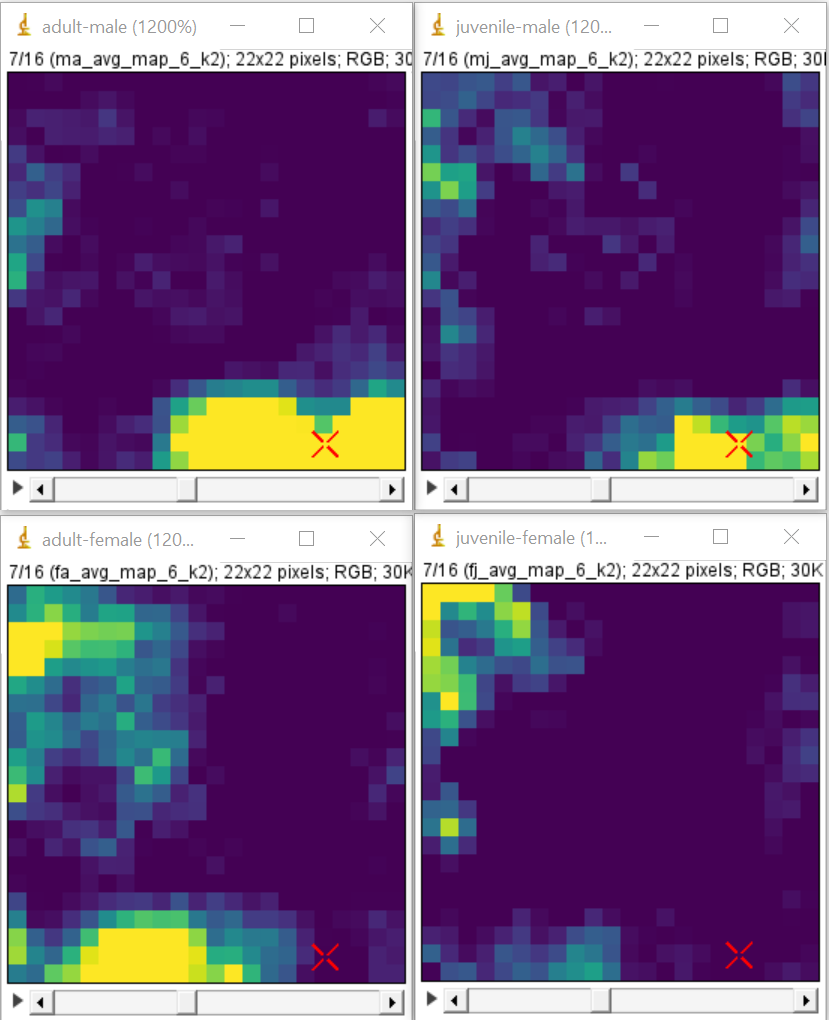}
        \caption{Output of channel 7 of second convolutional layer (9x9 kernel).}
        \label{fig:maps4}
    \end{subfigure}
     \begin{subfigure}{0.49\textwidth}
        \includegraphics[width=\textwidth]{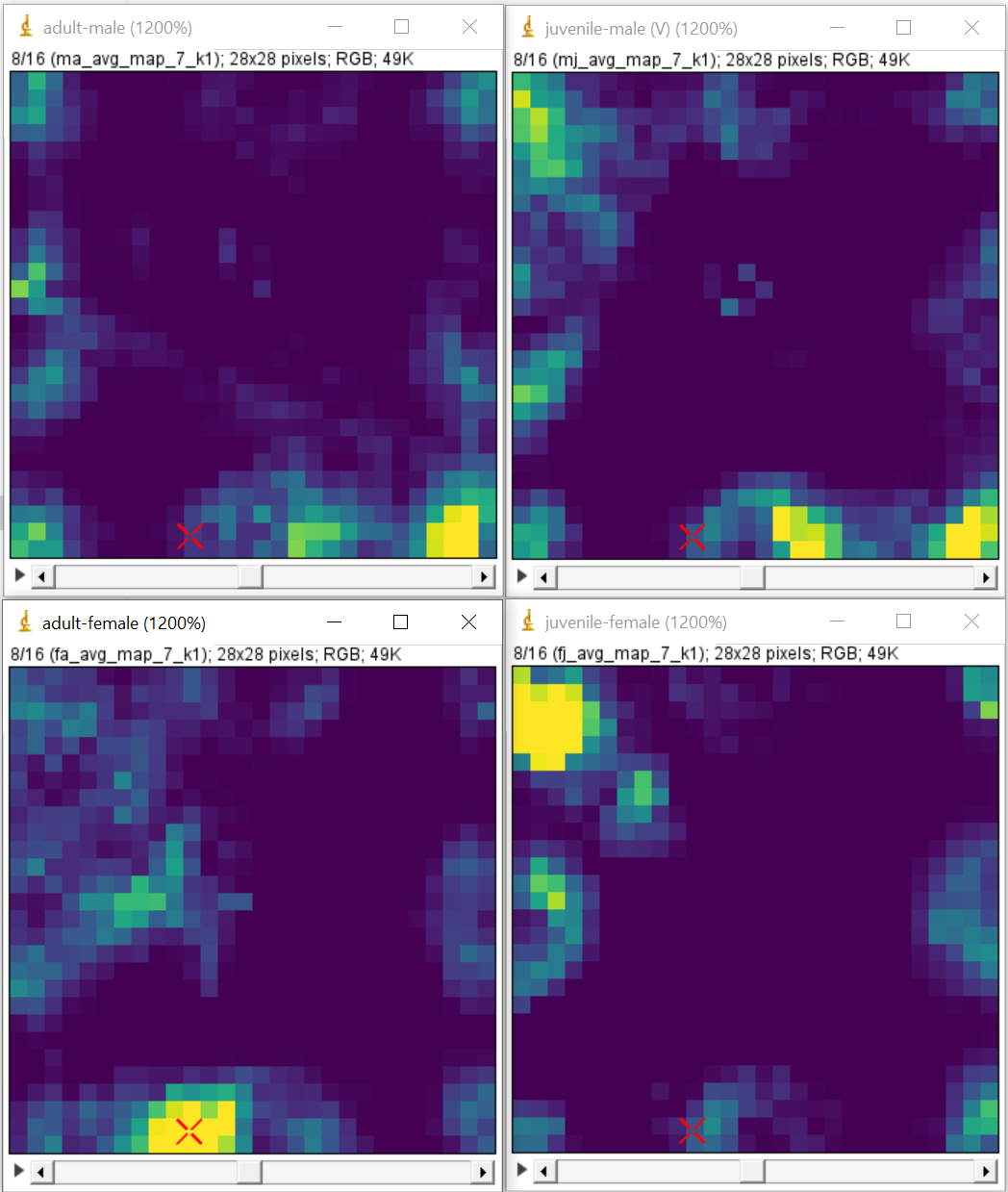}
        \caption{Output of channel 8 of first convolutional layer (3x3 kernel).}
        \label{fig:maps1}
    \end{subfigure}
    \hfill
    \begin{subfigure}{0.49\textwidth}
        \includegraphics[width=\textwidth]{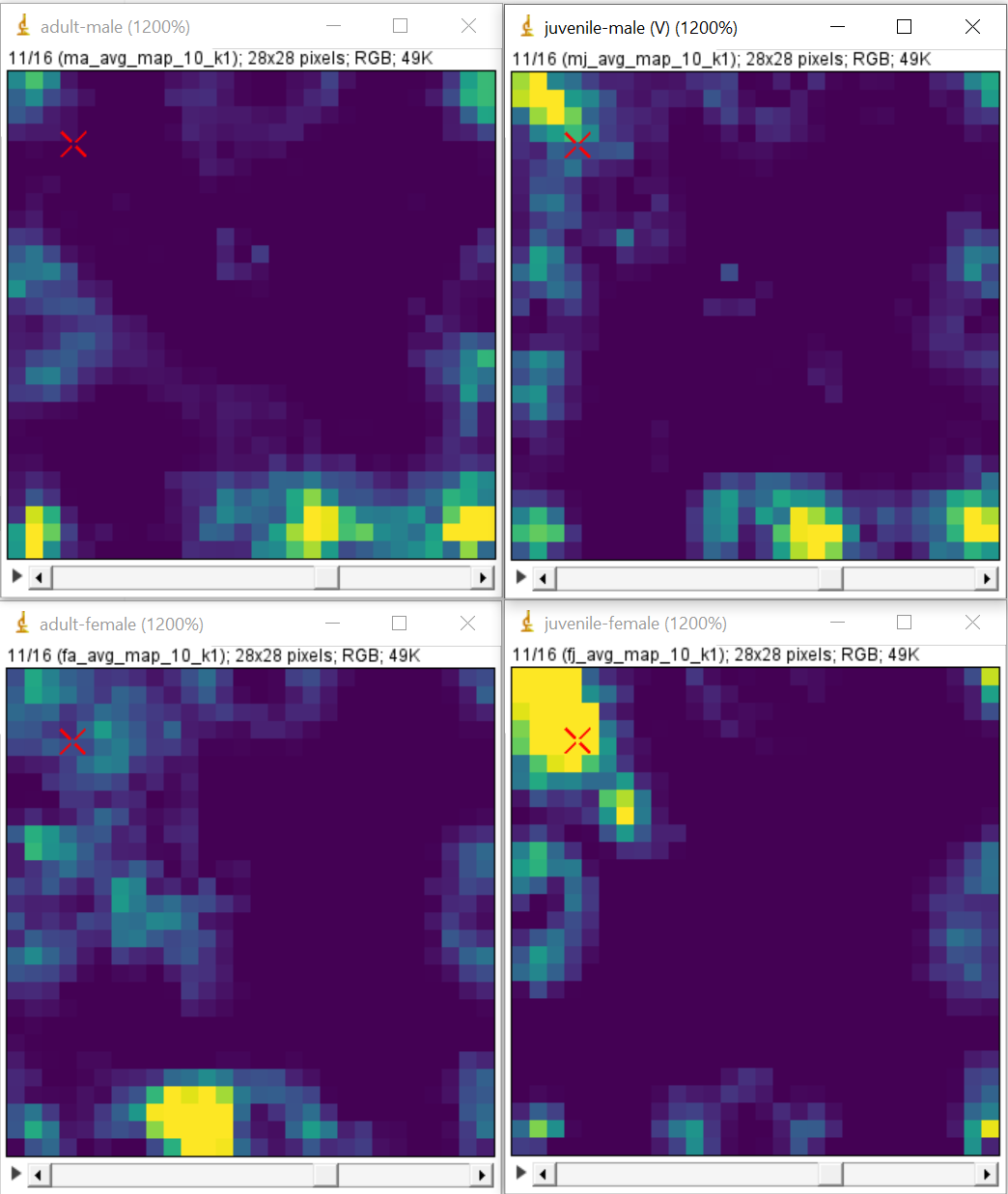}
        \caption{Output of channel 11 of first convolutional layer (3x3 kernel).}
        \label{fig:maps2}
    \end{subfigure}    
    \caption{Activation maps of a few channels of the convolution layers averaged over animals of the same class. Red crosses highlights the spots where we see considerable difference between activation maps. For each figure (a, b, c, d), AM is on top-left, JM on top-right, AF on bottom-left and JF on bottom-right.}
    \label{fig:maps}
\end{figure}

Given these results, we look at the activation maps of the convolution kernels in the network. In Figure \ref{fig:maps}, we see the average activation maps over all the individuals of each class for some of the 32 convolutional layers channels (16 each). We can see that there's actually no misclassification between \texttt{adult-males} and \texttt{adult-females} mice because kernels learned quickly how to distinguish them. In Figure \ref{fig:maps3} it is possible to observe the patterns related to the upper left corner of the cage, where the dome is placed, for male and female mice. Activation maps are exactly opposite one to another. While adult females have a strong predilection for the region where a dome was installed, the adult males seem to correlate much less with the same region. However, In Figure \ref{fig:maps1} an activation pattern change between \texttt{adult-females} and \texttt{juvenile-females} can be seen.

We also see that males correlate more with the lower part of the cage, especially between the lower feeder and the water spot.
Maps of Figure \ref{fig:maps4} for both \texttt{adult-males} and \texttt{juvenile-males} are similar, and response is particularly high in the lower border of the cage, which is also visible in Figures \ref{fig:maps1} and \ref{fig:maps2}. In the same figure it is evident that \texttt{adult-females} also show a similar pattern but slightly displaced to the left.  
It's worth noting that no relevant features are observed in \texttt{juvenile-males} with relation to other groups, in general, it has either the features of \texttt{adult-males} or \texttt{juvenile-female} (see figures \ref{fig:maps4} and \ref{fig:maps2}), which justify the strong level of misclassification in this class. This suggests that \texttt{juvenile-males} mice may take less time to mature, and that their stopping behavior choices at juvenile age are still not fully modulated by sexual differences.

\section{Conclusion}
\label{sec:conclusion}

In this work, we proposed a concatenation of single-layer convolutional neural network to identify the sex and age of animals based on stopping spots over time. Our work shows that, while females have more straightforward, identifiable features, stopping spots did not show distinctive features in males over age.

Observation of activation maps helped to understand which regions are more decisive in classification and may help in focusing behavioral analysis on what mice actually do in these regions. The initial cues indicate that nidification and feeding patterns are the ones that distinguish sex, while the age patterns are still a challenge for further work.

As next steps, we propose to keep investigating the stopping spots time series in the cage level, to determine which traits correlate with the hierarchy and main stereotypes of dominant/submissive individuals, as well as the variability of these traits with age and sex.

\bibliographystyle{splncs04}
\bibliography{main}

\end{document}